%% file: iser2020jjl.tex
\documentclass[a4paper]{styles/svmult}
\usepackage{mathptmx}       % selects Times Roman as basic font
\usepackage{helvet}         % selects Helvetica as sans-serif font
\usepackage{courier}        % selects Courier as typewriter font
\usepackage{type1cm}        % activate if the above 3 fonts are                           
\usepackage{subfigure}
\usepackage{makeidx}         % allows index generation
\usepackage{graphicx}        % standard LaTeX graphics tool
\usepackage{multicol}        % used for the two-column index
\usepackage[bottom]{footmisc}% places footnotes at page bottom
\usepackage[colorlinks]{hyperref}
\usepackage{color,xcolor}
\usepackage{caption}
\usepackage{multirow}
\usepackage{indentfirst}
\usepackage{amsmath}
\usepackage{bm}
\usepackage{physics}
\usepackage{amsfonts}
\usepackage{mathrsfs}

\graphicspath{{figures/}}
\DeclareGraphicsExtensions{.png,.jpg,.eps,.pdf}

\makeindex             % used for the subject index
\begin{document}

\title*{CMPCC: Corridor-based Model Predictive Contouring Control for Aggressive Drone Flight}

\author{Jialin Ji* \textsuperscript{1}, Xin Zhou* \textsuperscript{1}, Chao Xu \textsuperscript{1}, Fei Gao \textsuperscript{1}}

% \authorrunning{J. Ji, X. Zhou, C. Xu, F. Gao.}
% \titlerunning{Corridor based Model Predictive Contouring Control for Aggressive Drone Flight}
\institute{\textsuperscript{1} College of Control Science and Engineering, Zhejiang University. * Equal contributors.\\
E-mail: $\{$jlji, iszhouxin, cxu, and fgaoaa$\}$@zju.edu.cn\\
This work was supported by the Fundamental Research Funds for the Central Universities underGrant 2020QNA5013.
}

\maketitle

\vspace{-3cm}

\input{sections/1_introduction}

\input{sections/2_methodology}
\input{sections/3_experiments}
\input{sections/4_conclusion}

\bibliographystyle{plain}
\bibliography{iser2020jjl}
\end{document}

%% file: sections/1_introduction.tex
\section{Introduction}
Among the criteria of designing autonomous quadrotors, generating optimized trajectories and tracking the flight paths precisely are two critical components in the action aspect. 
As shown in our recent work \textit{Teach-Repeat-Replan}~\cite{gao2020teach}, a cascaded planning framework with global trajectory generation and local collision avoidance support agile flights under the user preferable routines. 
In work~\cite{gao2020teach}, though the first criteria is met by the global and local planners, the controller has no guarantees on tracking the generated motion precisely. 
Also, in industrial applications, the planner and controller of a quadrotor are mostly independently designed, making it hard to tune the joint performance in different applications.

Some works~\cite{seo2019robust, li2020fast} attempt to compensate uncertainties introduced by disturbances, by designing error-tolerated trajectory planning methods based on Hamilton-Jacobi Reachability Analysis~\cite{bansal2017hamilton}. 
These works set handcrafted disturbance bounds, making it too conservative to find a feasible solution among dense obstacles. 
\textit{Tal et.al, }~\cite{tal2018accurate} propose control systems for accurate trajectory tracking that improves tracking accuracy.
Nevertheless, they still try to track the unreachable trajectory when facing violent disturbance instead of adjusting the primary trajectory. 
If un-negligible disturbance occurs, local replanners such as~\cite{zhou2019robust, usenko2017real} can plan motions to rejoin the reference quickly, but they are inferior to give a proper temporal distribution. 
The closest work to this paper~\cite{liniger2015optimization} applies Model predictive contouring control (MPCC)~\cite{lam2010model} as the planner for miniature car racing, where safety constraints are established by modeling linear functions from the boundary of the racing track. 
However, such constraints are not directly available for a quadrotor in unstructured environments.
What's more, due to the limited planning horizon, MPCC cannot guarantee feasibility and heavily relies on proper parameter tuning.

To bridge this gap, we propose an efficient, receding horizon, local adaptive low-level planner as the middle layer between our original planner and controller. 
Our method is named as corridor-based model predictive contouring control (CMPCC) since it builds upon on MPCC ~\cite{lam2010model} and utilizes the flight corridor as hard safety constraints. 
It optimizes the flight aggressiveness and tracking accuracy simultaneously, thus improving our system's robustness by overcoming unmeasured disturbances. 
Our method features its online flight speed optimization, strict safety and feasibility, and real-time performance and it is released as a low-level plugin for a large variety of quadrotor systems.
We summarize our contribution as follows:
\begin{enumerate}
  \item We propose an efficient and disturbance-adaptive receding horizon low-level planner, which generates a collision-free and dynamic feasibile trajectory with adjusted temporal allocation in real time.
  \item We propose a method of building appropriate constraints by constructing a forward spanning polygon tube from corresponding polyhedron and setting a terminal velocity from reference trajectory.
  \item We integrate the proposed methods into a fully autonomous quadrotor system, and release our software for the reference of the community\footnote{\url{https://github.com/ZJU-FAST-Lab/CMPCC}}.
\end{enumerate}

%% file: sections/2_methodology.tex
\section{Methodology}
% \vspace{-1em}

\begin{figure}[ht]
    \centering
    {\includegraphics[height=0.6\columnwidth]{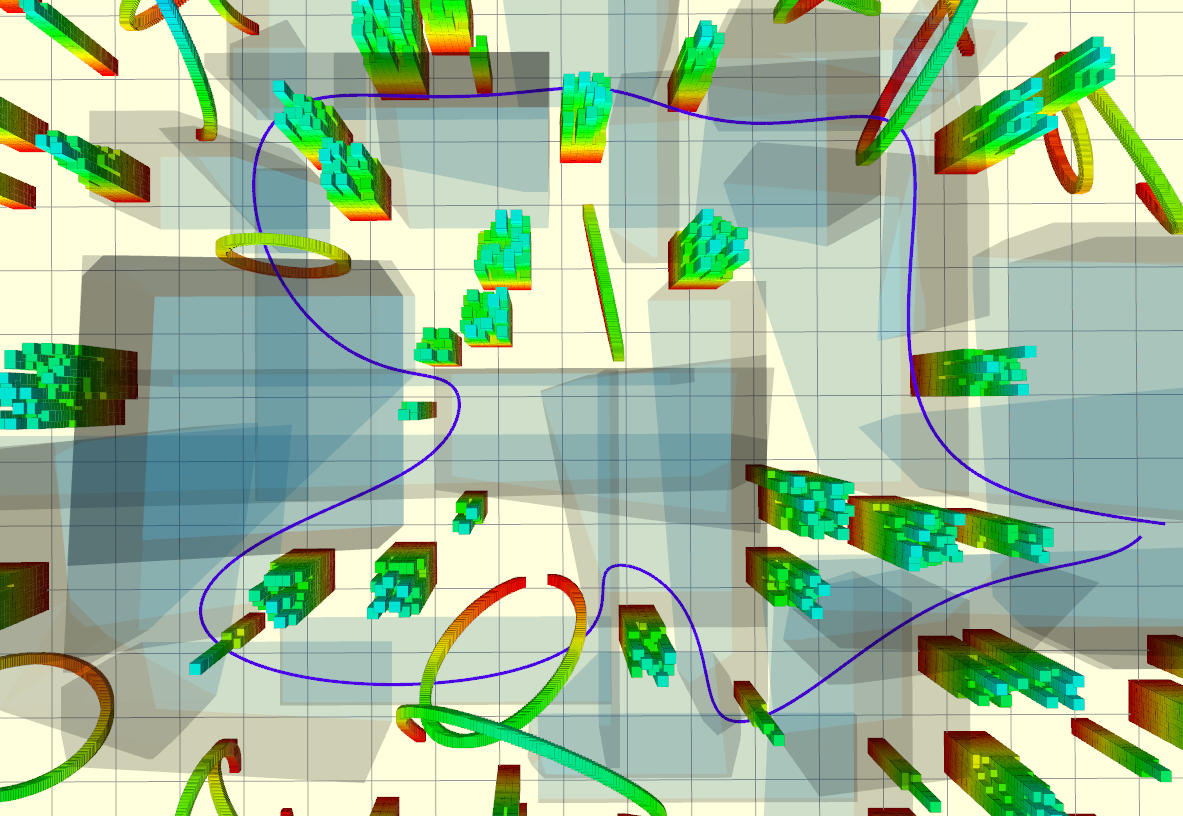}}
    \caption{Global reference trajectory and Flight corridors}
    \label{fig:flight-corridors}
\end{figure}

We get the global optimized reference trajectory and the flight corridor from our previous work \textit{Teach-Repeat-Replan}~\cite{gao2020teach}, as shown in Fig.~\ref{fig:flight-corridors}.

The global trajectory $p(t)$ is an optimized smooth curve in the space, parameterized by its original time $t$, given by
\begin{eqnarray}
p = [x_p,y_p,z_p]^T, p(t): \left[t_0, t_m \right] \rightarrow \mathbb{R}^3.
\end{eqnarray}

\begin{figure}[ht]
    \centering
    \subfigure[\label{fig:retiming}]
    {\includegraphics[height=0.25\columnwidth]{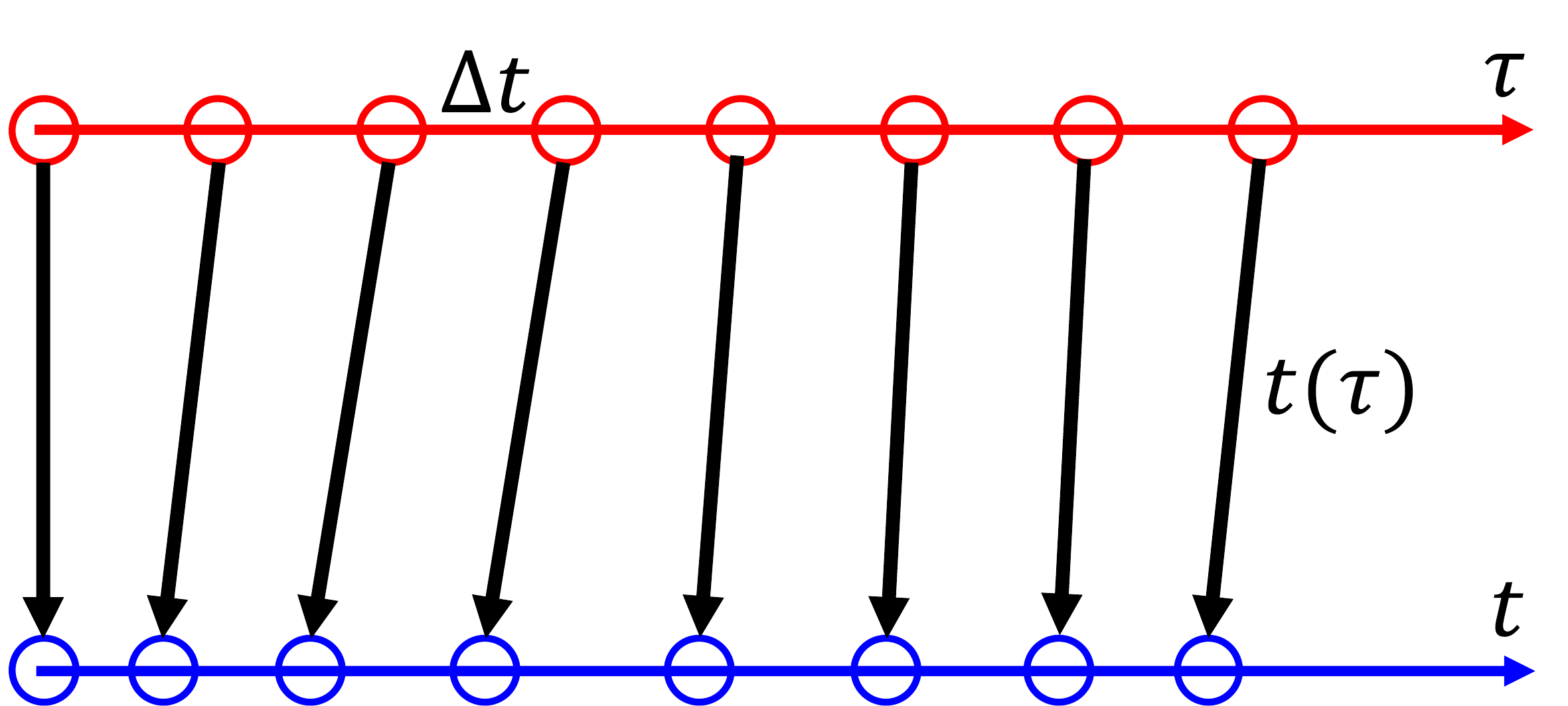}}
    \subfigure[\label{fig:obj}]
    {\includegraphics[height=0.25\columnwidth]{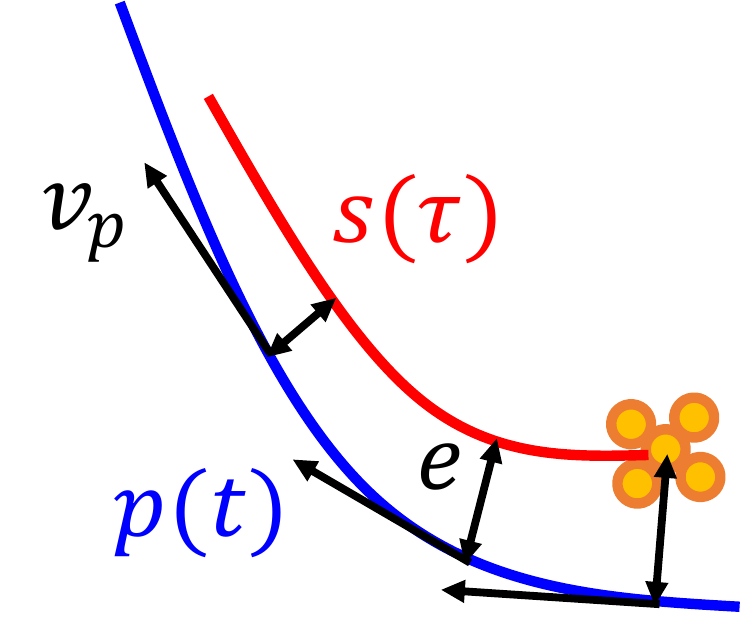}}
    \caption{ (a) Re-timing function (b) Components of the objective of CMPCC}
\end{figure}

We design a re-timing function $\{t(\tau) : \tau \rightarrow t\}$ to map the original time variable $t$ to a new time variable $\tau$, shown in Fig.~\ref{fig:retiming}. And we also design the adjusted local trajectory $s(\tau)$, given by
\begin{eqnarray}
s=[x,y,z]^T,s(\tau):[0,\tau_m]\rightarrow \mathbb{R}^3, 
\end{eqnarray}
where $\tau_m$ is the duration of predictive horizon. 
The optimization objective is shown in Fig.~\ref{fig:obj}, where $v_p = \dot p(t(\tau))$ indicates the speed of the point on global reference trajectory after re-timing, and the tracking error $e(\tau) = \left\|s(\tau) - p(t(\tau))\right\|$.
The objective trades off the minimization of $\{\int_0^{\tau_m} e(\tau) d\tau\}$ and the maximization of $\{\int_0^{\tau_m} v_p d\tau\}$ by optimizing $t(\tau)$ and $s(\tau)$.

Note that $v_p = \dot p(t(\tau)) = p'(t)\cdot \dot{t}(\tau)$, where $p'(t)$ is definite according to the global trajectory $p(t)$ but $\dot{t}(\tau)$ truly indicates the traveling progress, i.e. $\dot{t}(\tau) > 1$ means the drone travels faster than and $\dot{t}(\tau) < 1$ means the opposing situation. Thus the objective indicating the tradeoff of tracking error $\left\|s(\tau) - p(t(\tau))\right\|$ and traveling progress 
$\dot{t}(\tau)$ is given by
\begin{eqnarray}
J = \int_{0}^{\tau_m} \left( \left\| s - p \right\|^2- \rho\dot t\right) d\tau, \label{(0)}
\end{eqnarray}
where $\rho$ is a weight of the aggressiveness. 
We model the system as a $3^{rd}$-order integral model and mark $v_\mu:=\dot \mu(\tau)$, $a_\mu:=\ddot \mu(\tau)$, $j_\mu:=\dddot \mu(\tau), \mu = {x,y,z,t}$. Then the states and inputs of the system is given by
\begin{eqnarray}
    &\mathbf x = [x,v_x,a_x,y,v_y,a_y,z,v_z,a_z,t,v_t,a_t]^T, \label{(1)}\\
    &\mathbf u = [j_x, j_y, j_z, j_t]^T. \label{(2)}
\end{eqnarray}

They are discretized by $\Delta t$ with the length of predictive horizon $N$, and the optimization problem turns into a receding horizon MPC with  linear state-transfer equations given by
\begin{eqnarray}
\mathbf x^{(k+1)} = \mathbf{A_d} \mathbf x^{(k)} + \mathbf{B_d} \mathbf u^{(k)} \label{(state-transfer-equation)}
\end{eqnarray}
at the $k$-th time-step, 
where $\mathbf{A_d}$ and $\mathbf{B_d}$ are given by
\begin{eqnarray}
    \mathbf{A_d} = \bigoplus_{i=1}^4 
        \begin{pmatrix}
            1 & \Delta t & \frac{1}{2}\Delta t^2 \\
            0 & 1 & \Delta t \\
            0 & 0 & 1 \\
        \end{pmatrix},
    \mathbf{B_d} = \bigoplus_{i=1}^4 
        \begin{pmatrix}
            0\\
            0\\
            \Delta t\\
        \end{pmatrix},
\end{eqnarray}
and the objective \ref{(0)} is discretized as
\begin{eqnarray}
J = \sum\limits_{k=1}^{N} \left\{\sum\limits_{\mu=x,y,z} {\left(\mu^{(k)}-\mu_p(t^{(k)})\right)^2} - \rho\cdot v_t^{(k)}\right\}, \label{(nonlinear)}
\end{eqnarray}
which is a nonlinear function of $\mathbf x$ and $\mathbf u$. Thanks to the framework of receding horizon, we linearize $\mu_p(t^{(k)})$ by
\begin{eqnarray}
\mu_p(t^{(k)}) = \mu_p(\theta^{(k)}) + \mu_p'(\theta^{(k)}) \cdot \left(t^{(k)} - \theta^{(k)}\right),
\end{eqnarray}
where $\theta^{(k)}$ is the optimized result of $t^{(k+1)}$ in the last horizon.
Thus the objective \ref{(nonlinear)} can be represented as quadratic, given by
\begin{eqnarray}
J = \sum\limits_{k=1}^{N} \left\{
    {\mathbf{x}^{(k)}}^T S_1^T Q_k S_1 \mathbf{x}^{(k)} + q_k^T S_2 \mathbf{x}^{(k)}
\right\},
\end{eqnarray}
where $S_1$ and $S_2$ are selection matrices such that
\begin{eqnarray}
\begin{pmatrix}
    x^{(k)}, y^{(k)}, z^{(k)}, t^{(k)}
\end{pmatrix}^T = S_1 \mathbf{x}^{(k)}, \\
\begin{pmatrix}
    x^{(k)}, y^{(k)}, z^{(k)}, t^{(k)}, v_t^{(k)}
\end{pmatrix}^T = S_2 \mathbf{x}^{(k)},
\end{eqnarray}
and 
\begin{eqnarray}
& Q_k =
\begin{pmatrix}
    1 & 0 & 0 & -x'_p(\theta^{(k)}) \\
    0 & 1 & 0 & -y'_p(\theta^{(k)}) \\
    0 & 0 & 1 & -z'_p(\theta^{(k)}) \\
    -x'_p(\theta^{(k)}) & -y'_p(\theta^{(k)}) & -z'_p(\theta^{(k)}) & \sum\limits_{\mu=x,y,z} {\mu'_p(\theta^{(k)})}^2 \\
\end{pmatrix}, \\
& q_k =
\begin{pmatrix}
    2c^{(k)}_x, 2c^{(k)}_y, 2c^{(k)}_z, \sum\limits_{\mu=x,y,z} \left\{-2\mu'_p(\theta^{(k)})c^{(k)}_\mu\right\}, -\rho
\end{pmatrix}^T, \\
& c^{(k)}_\mu =
\mu'_p(\theta^{(k)})\cdot \theta^{(k)} - \mu_p(\theta^{(k)}), \mu={x,y,z}.
\end{eqnarray}

\begin{figure}[ht]
\centering
    \subfigure[\label{fig:ft1}]
{\includegraphics[height=0.32\columnwidth]{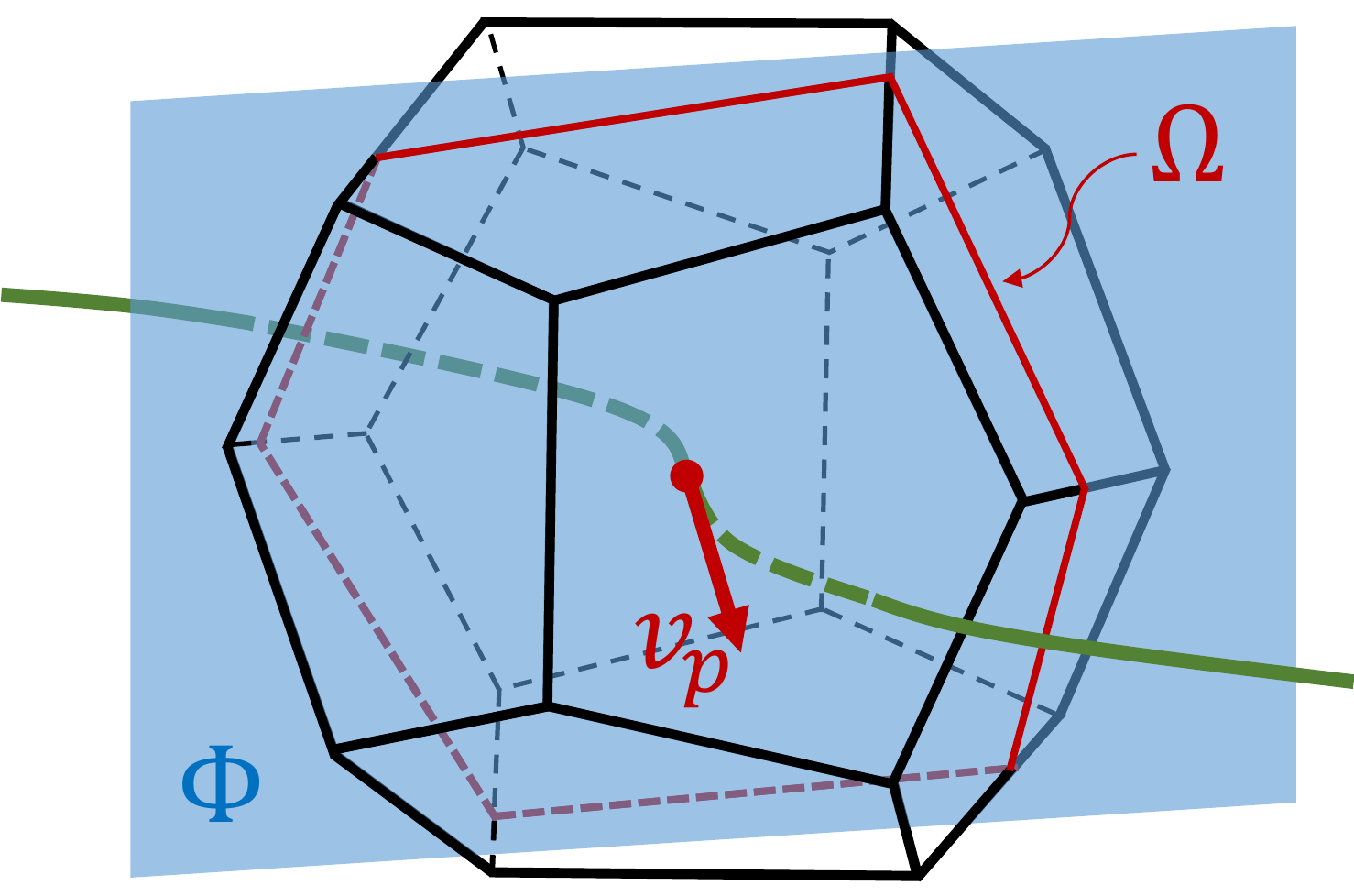}}
    \subfigure[\label{fig:ft2}]
{\includegraphics[height=0.32\columnwidth]{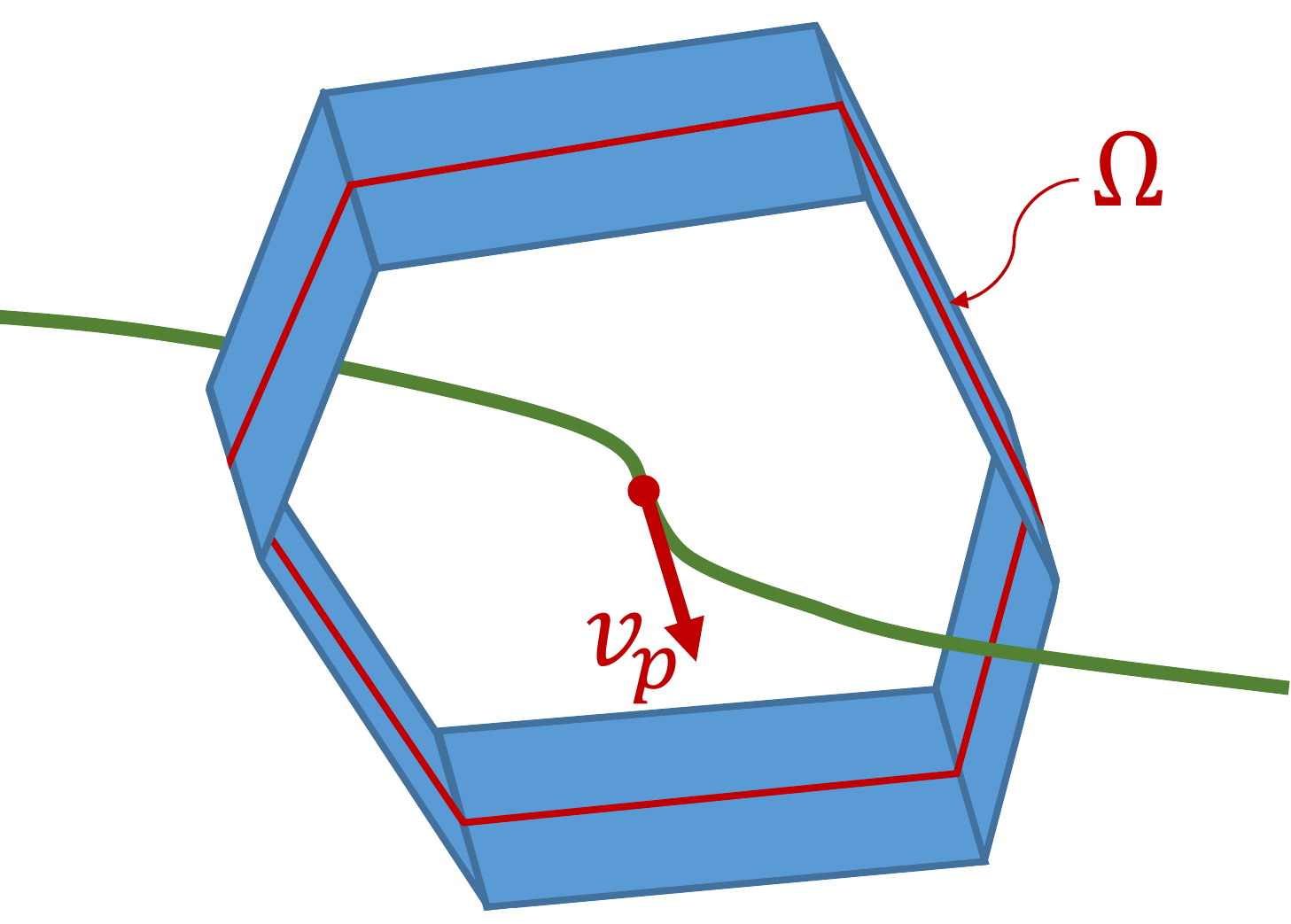}}
\caption{Generation of polygon tube}
\end{figure}

\begin{figure}[ht]
    \centering
    {\includegraphics[height=0.35\columnwidth]{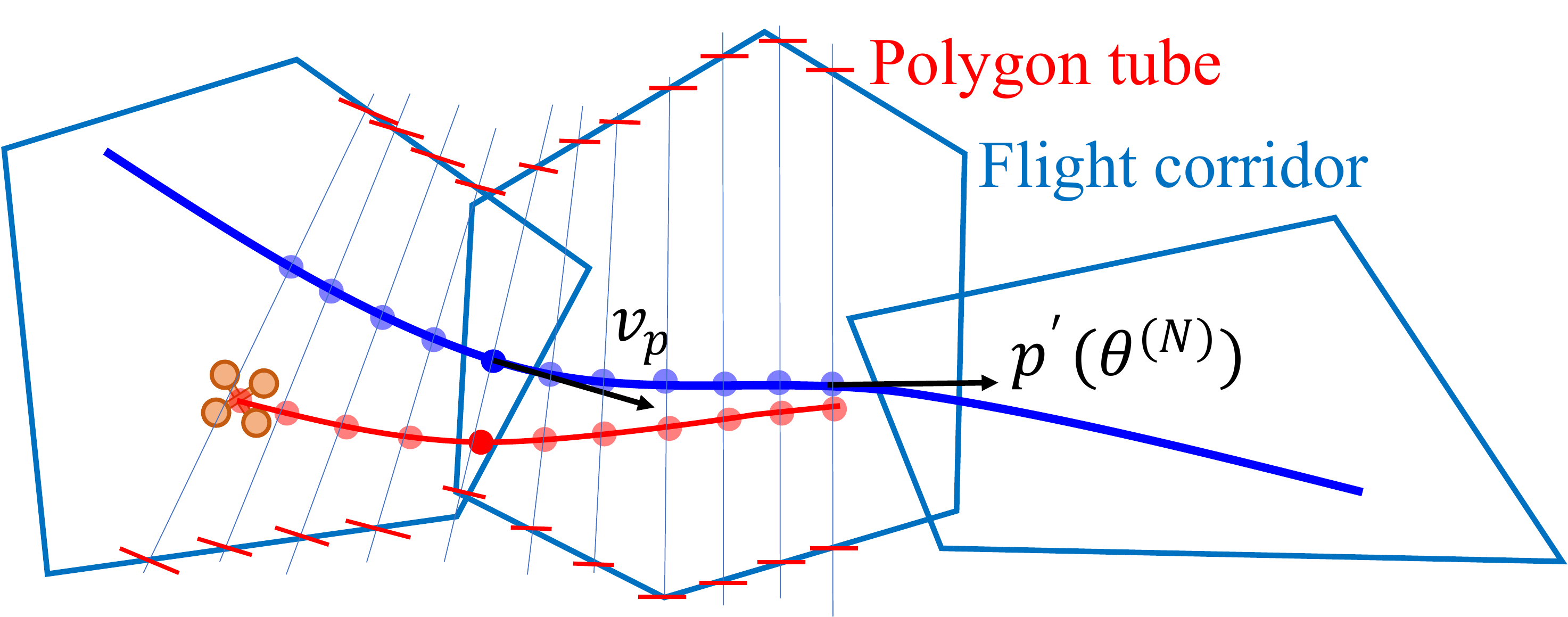}}
    \caption{Sequence of linear inequality constraints for safety}
    \label{fig:sq}
\end{figure}

Then we construct linear inequality constraints. 
For a given reference point $p(\theta^{(k)})$ on the global trajectory, we define $\rm\Omega$ as the intersection of $v_p$'s normal plane $\rm\Phi$ with the corresponding polyhedron.
As shown in Fig.~\ref{fig:ft1}, the resulting $\rm\Omega$ is a convex polygon. 
Then each edge of $\rm\Omega$ expands a plane sweeping along the direction of $v_p^{(k)}$, which gives a polygon tube, as shown in Fig.~\ref{fig:ft2}.
The inner side of this tube is considered as the safe space near $p(\theta^{(k)})$ and will be modeled as inequality constraints:
\begin{eqnarray}
    \mathbf C^{(k)} \cdot [x^{(k)}, y^{(k)}, z^{(k)}]^T \le \mathbf b^{(k)}, k=1,2,3,...,N.
\end{eqnarray}
The sequence of the polygon tubes is shown in Fig.~\ref{fig:sq}.

In order to guarantee the dynamic feasibility, physical limits for each state are set as inequality constraints:
\begin{eqnarray}
    l_x \le S_3 \mathbf{x}^{(k)} \le  u_x, \\
    l_u \le \mathbf{u}^{(k)} \le  u_u,
\end{eqnarray}
where $l_x,l_u,u_x,u_u$ are lower and upper bounds of velocity, acceleration and jerk at each time step. $S_3$ is another selection matrix such that 
\begin{eqnarray}
    [v_x^{(k)},a_x^{(k)},v_y^{(k)},a_y^{(k)},v_z^{(k)},a_z^{(k)}]^T = S_3 \mathbf{x}^{(k)}
\end{eqnarray}

In addition to physical limits for each state, a terminal velocity constraint is added such that the terminal speed should be less than $p'(\theta^{(N)})$ in the predictive horizon.
Thus, feasibility can be guaranteed since the reference trajectory is globally optimal.

The optimization problem is formulated as a standard quadratic program (QP):

\begin{eqnarray}
\min\limits_{\mathbf{\bar x}} & \mathbf{\bar x}^T Q \mathbf{\bar x} + q^T \mathbf{\bar x} \\
\mathbf{s.t.} & P\mathbf{\bar x} = h \\
& C \mathbf{\bar x} \le b 
\end{eqnarray}
where 
\begin{eqnarray}
    \mathbf{\bar x} = [{\mathbf x^{(1)}}; {\mathbf u^{(1)}}; {\mathbf x^{(2)}}; {\mathbf u^{(2)}}; ...; {\mathbf x^{(N)}}; {\mathbf u^{(N)}}]
\end{eqnarray}
and it is solved by OSQP\cite{osqp} with warm start speed-up.
In practice, we choose a $1s$ predictive horizon and the sampling interval $\Delta t = 0.05s$, which means $N = 20$. 
The performance of our algorithm is tested on an Intel i7-6700 CPU, with the average solving time around 5$ms$.

%% file: sections/3_experiments.tex
\section{Experiments}
\subsection{Experiments configuration}

\begin{figure}[ht]
    \centering
    {\includegraphics[height=0.45\columnwidth]{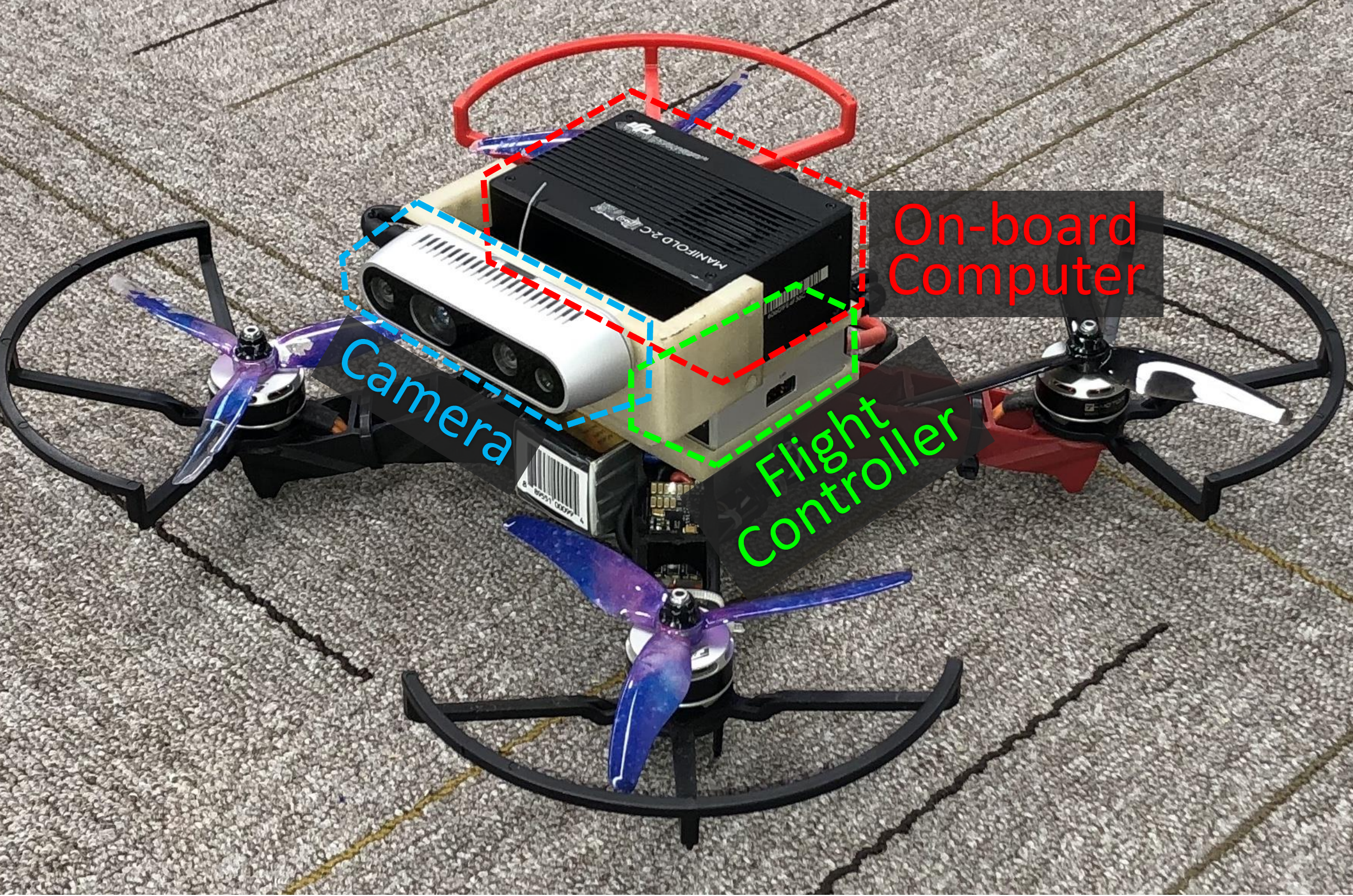}}
    \caption{The hardware of the UAV}
    \label{fig:hard}
\end{figure}
\begin{figure}[ht]
    \centering
    {\includegraphics[height=0.45\columnwidth]{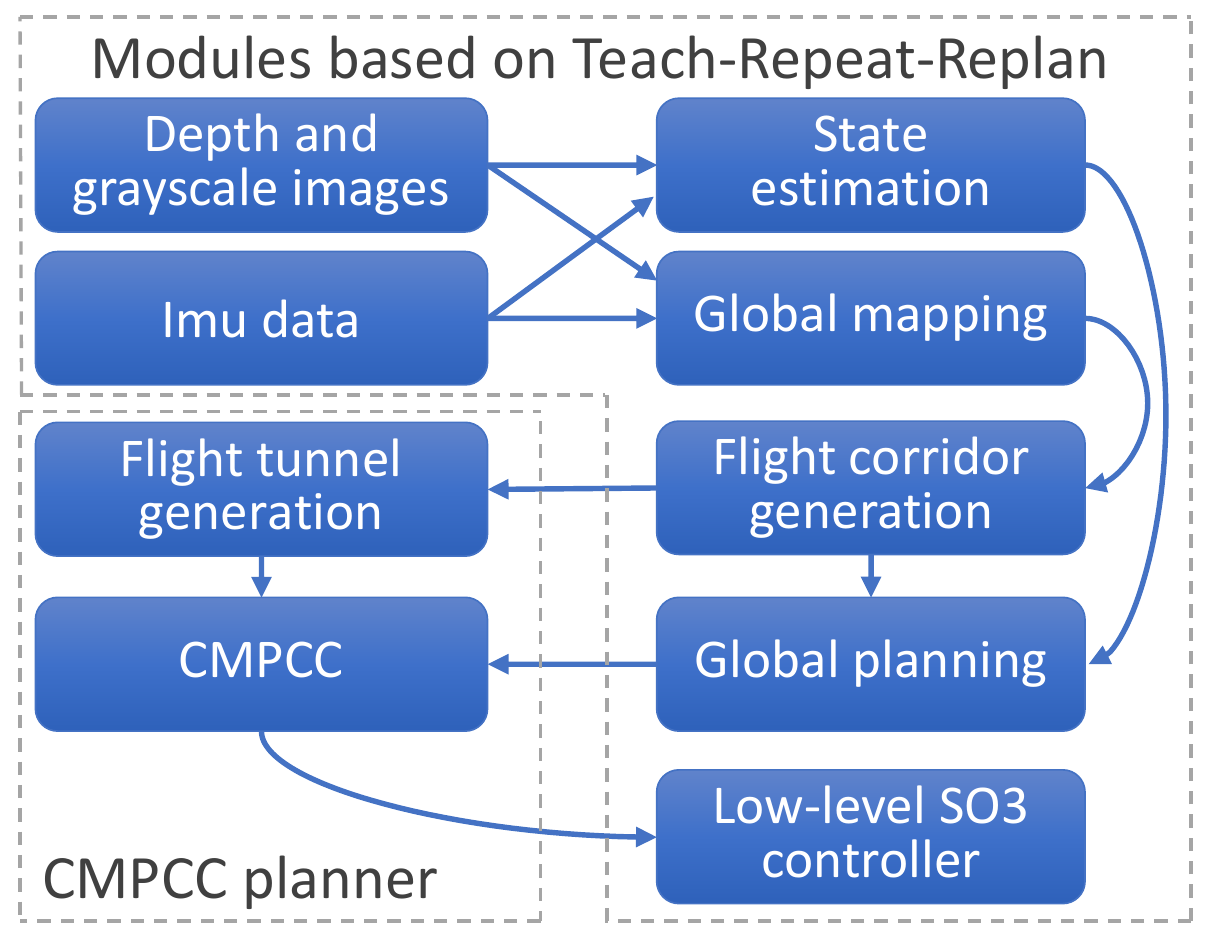}}
    \caption{The Software architecture of the UAV}
    \label{fig:soft}
\end{figure}

We use a self-developed quadrotor with an Intel Realsense D435i stereo camera\footnote{https://www.intelrealsense.com/depth-camera-d435i/} and a DJI N3 flight controller\footnote{https://www.dji.com/n3} for state estimation. 
All modules run solely on a DJI Manifold 2-C onboard computer \footnote{https://www.dji.com/manifold-2}.
Our system inherits the localization, mapping and global planning from the \textit{Teach-Repeat-Replan} system~\cite{gao2020teach}, where readers can check these modules in detail.
The overall hardware and software architecture of our system is shown in Fig.~\ref{fig:hard} and Fig.~\ref{fig:soft}.

\subsection{Autonomous flight with contact disturbance}

\vspace{-0.5cm}
\begin{figure}[ht]
    \centering
        \subfigure[\label{fig:disturbance1}]
    {\includegraphics[height=0.32\columnwidth]{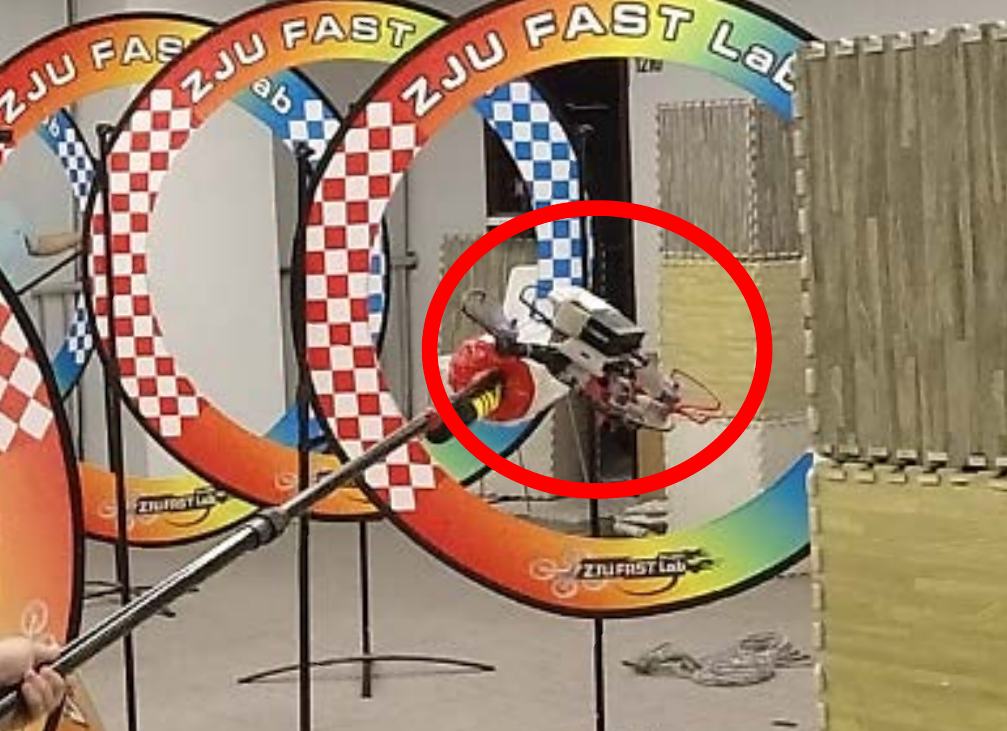}}
        \subfigure[\label{fig:disturbance2}]
    {\includegraphics[height=0.32\columnwidth]{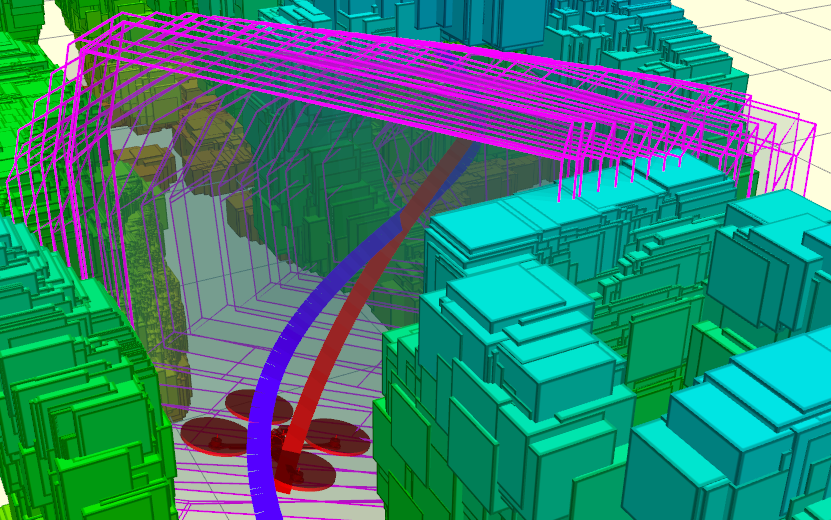}}
    \caption{Circumstance of instant force disturbance}
\end{figure}

\begin{figure}[ht]
    \centering
    {\includegraphics[width=0.65\columnwidth]{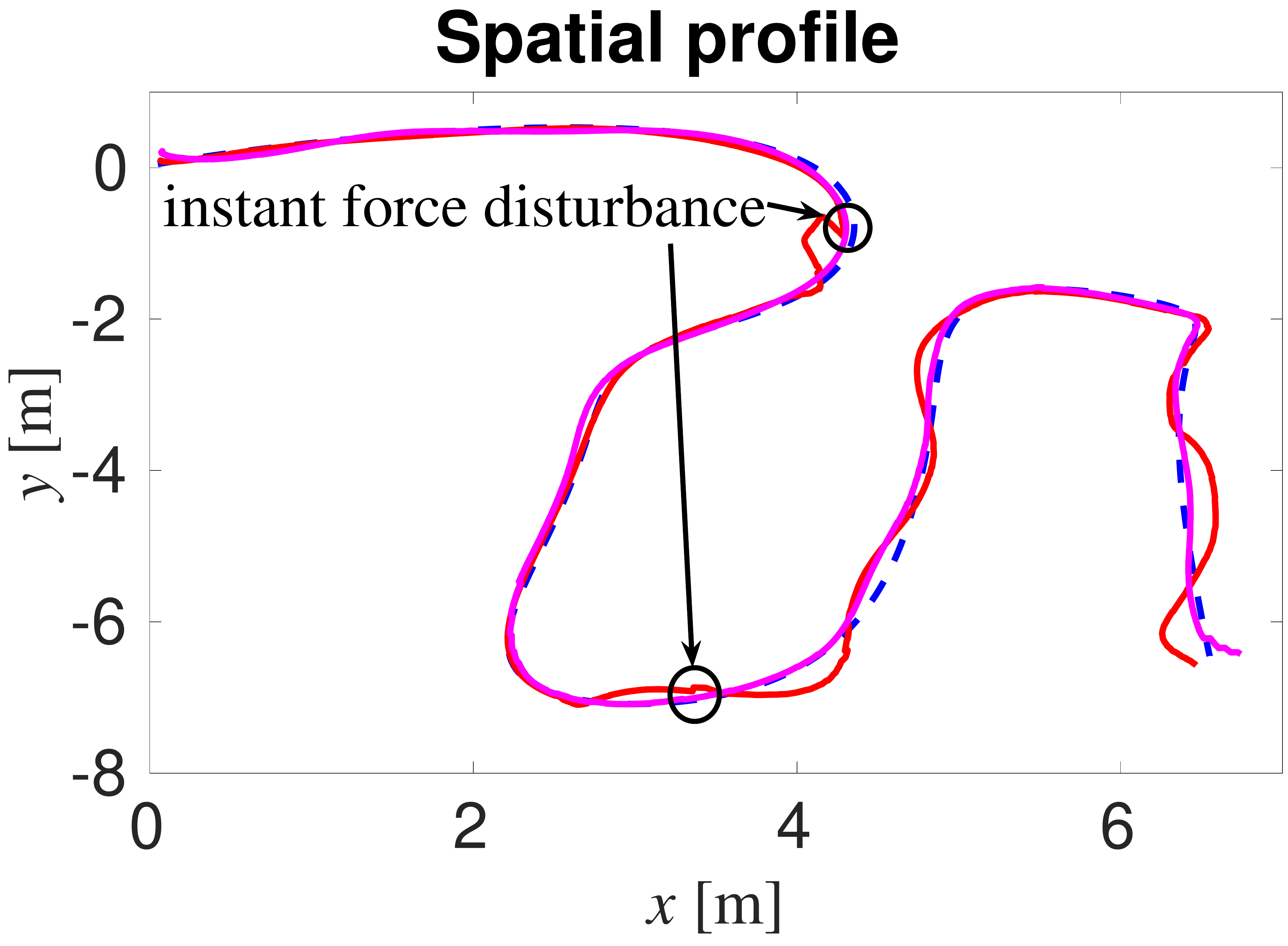}}
    \caption{Spatial profile facing instant force disturbance}
    \label{fig:dis-l}
\end{figure}

\begin{figure}[ht]
    \centering
    {\includegraphics[width=0.65\columnwidth]{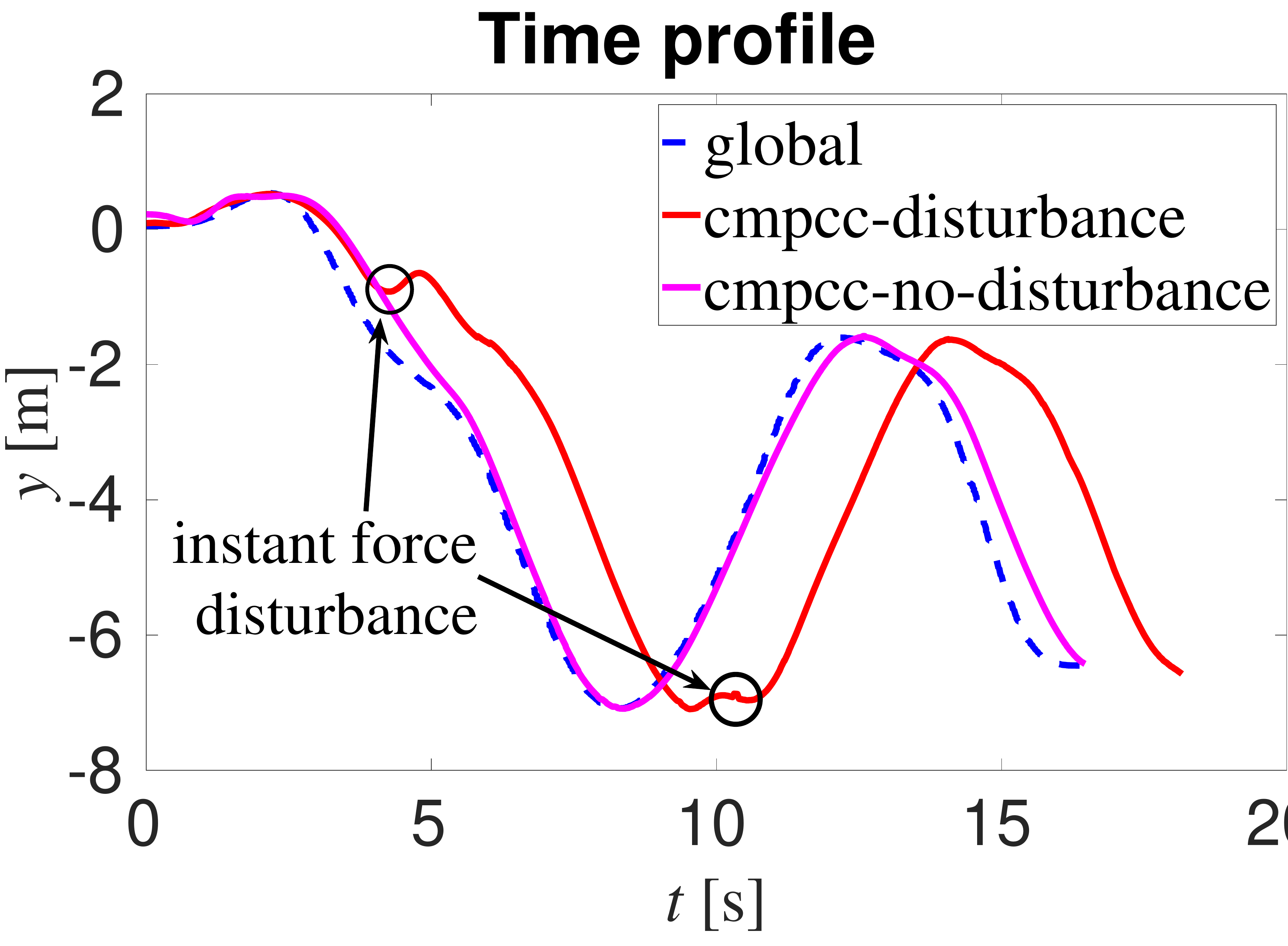}}
    \caption{Time profile facing instant force disturbance}
    \label{fig:dis-r}
\end{figure}

We apply challenging force disturbance to the drone to test the performance of the proposed CMPCC, as shown in Fig.~\ref{fig:disturbance1}. 
The global trajectory (blue), our locally re-planned trajectory (red), and the geometry constraints (magenta) after the hit are visualized in Fig.~\ref{fig:disturbance2}. 
As seen in the top-down view of the experiment in Fig.~\ref{fig:dis-l}, oscillation occurs near the hit position, but the local trajectory soon converges to the global one. 
Also, as shown in Fig.~\ref{fig:dis-r}, the instant force disturbance changes the temporal distribution of the optimal trajectory, resulting in the delaying of the trajectory of cmpcc (red) relative to which without disturbance (magenta) and the global trajectory (blue).

% \vspace{-0.5em}
\subsection{Autonomous flight with wind disturbance}
\begin{figure}[ht]
    \centering
        \subfigure[\label{fig:fan}]
    {\includegraphics[height=0.42\columnwidth]{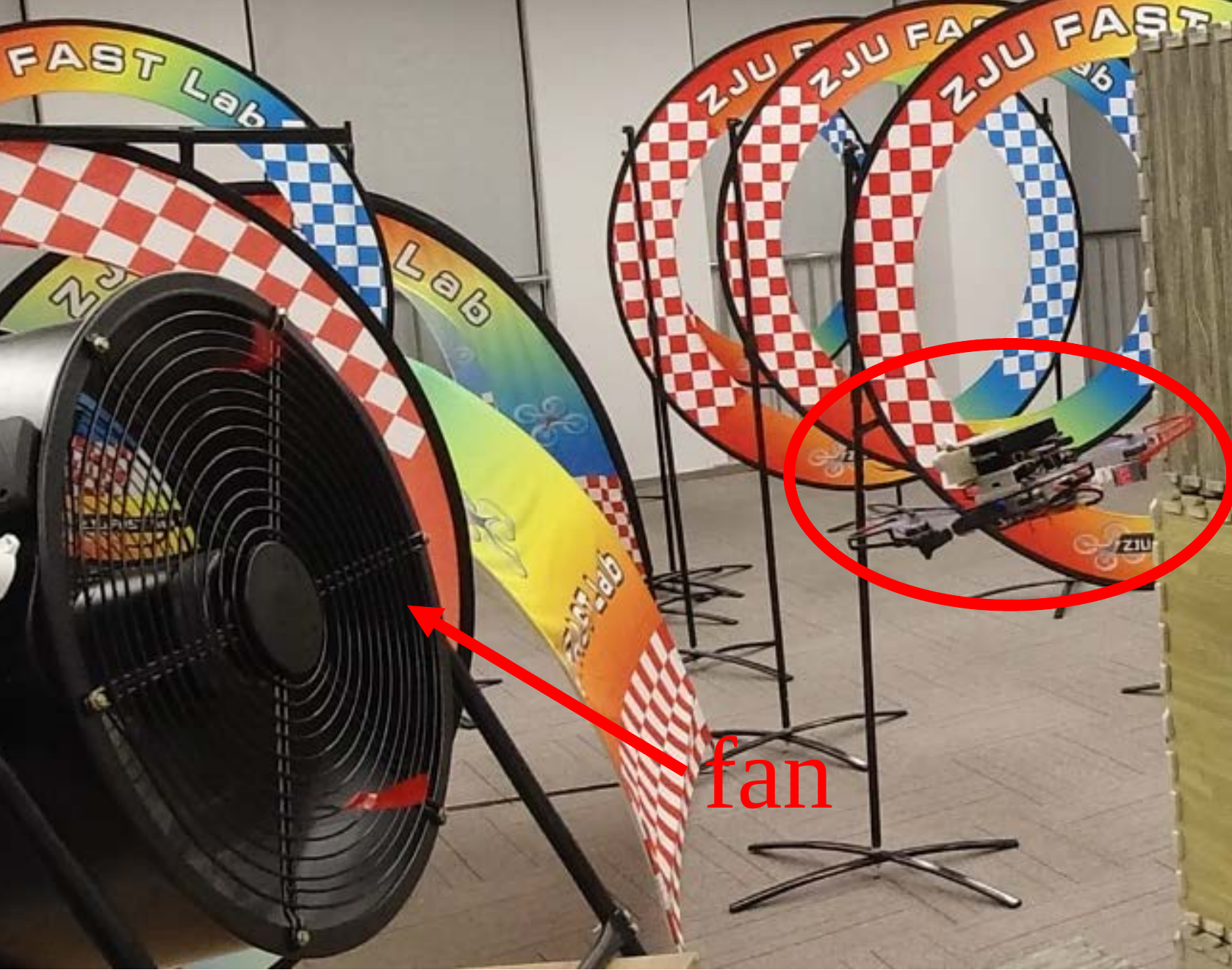}}
        \subfigure[\label{fig:fan-plot}]
    {\includegraphics[height=0.42\columnwidth]{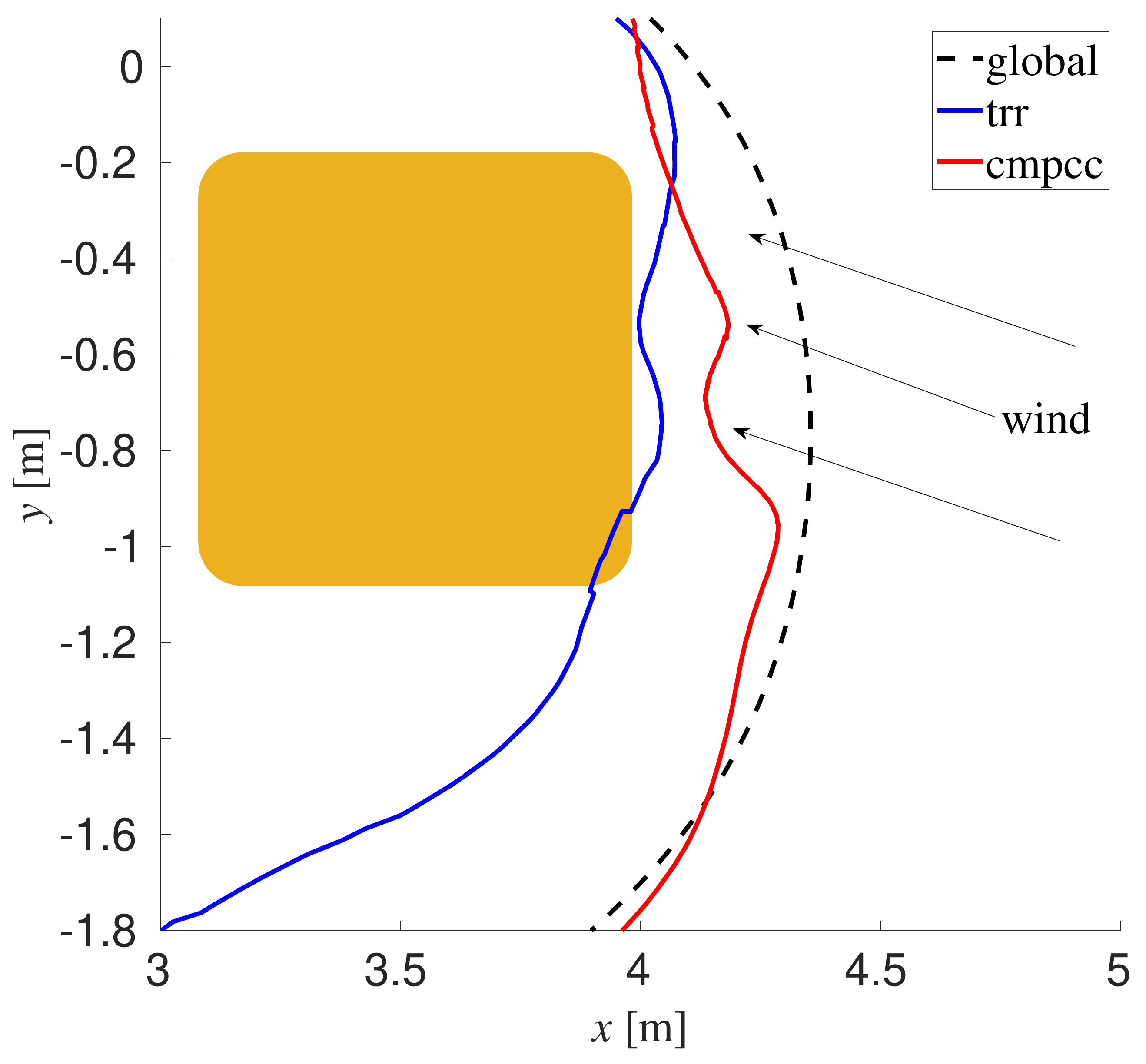}}
    \caption{The circumstance of wind disturbance}
\end{figure}

% \vspace{-0.5em}
We also test our method with wind disturbance by a fan, as shown in Fig.~\ref{fig:fan}. 
Without the proposed CMPCC, the quadrotor tracks the global trajectory generated by \textit{Teach-Repeat-Replan} with only a feedback controller, and collides with the nearby obstacle soon.
However, thanks to the safety guarantee, the proposed CMPCC re-plans a safe trajectory and rejoins the global reference quickly under the wind disturbance, as shown in Fig.~\ref{fig:fan-plot}.

The video of the experiments is available. \footnote{\url{https://www.youtube.com/watch?v=\_7CzBh-0wQ0}}

%% file: sections/4_conclusion.tex
% \vspace{-1em}
\section{Main Experimental Insights}

In practice, the flight performance of a quadrotor can be affected by many factors. 
Among all issues, the unexpected and unmeasurable disturbance is always an essential one for quadrotor autonomous navigation, especially for fast and aggressive flight. 
Recently, most autonomous quadrotor systems~\cite{oleynikova2020open, gao2020teach} are developed with several independent modules include controller, planner, and perception, with the assumption that a properly designed, smooth, derivative bounded trajectory can be tracked by a controller within high bandwidth.
However, this assumption does not always hold.
No matter how robust the feedback controller is, it's noted that it may fail when encountering drastic disturbance, such as immediate contact and a gust of wind, which are demonstrated in our experiments. 
Traditionally, people have to spend tons of time tuning the parameters of the feedback controller until a satisfactory performance.
In this work, as validated by our challenging experiments, the proposed intermediate low-level replanner successfully compensates disturbances by planning local safe trajectories and automatically adjusting the flight aggressiveness. 
Therefore, the robustness of fast autonomous flight is improved significantly. 
Moreover, thanks to the convex formulation, the proposed CMPCC is solved within 5 $ms$, which suits onboard usage well.

In experiments, we also observe that the polygon tube now we use heavily depends on the static corridor. Therefore it cannot handle the variation of the environment or dynamic obstacles. 
In the future, we plan to investigate the way to generate safety constraints for CMPCC online.